\documentclass[10pt,twocolumn,letterpaper]{article}

\usepackage[pagenumbers]{wacv} 

\definecolor{wacvcolor}{rgb}{0,0,0}

\usepackage{algorithm}
\usepackage{algpseudocode}
\usepackage{tikz}
\usetikzlibrary{arrows.meta,calc,positioning}

\usepackage{algorithm}
\usepackage{algpseudocode}
\usepackage{multirow}
\definecolor{wacvblue}{rgb}{0.21,0.49,0.74}
\usepackage[pagebackref,breaklinks,colorlinks,allcolors=wacvblue]{hyperref}

\title{A Sparse Topometric Graph \\ for Scalable Cross-Session Indoor Navigation}

\author{
Eun Sun Lee \quad
Minkyun Seo \quad
Gwangtak Bae \quad
Junho Kim \\
Seunggu Kang \quad
Jaesik Park$^{*}$ \quad
Young Min Kim$^{*}$ \\[6pt]
Seoul National University \\[2pt]
{\tt\small
\{eunsunlee,funboy0804,tak3452,82magnolia,jrkangsg,%
jaesik.park,youngmin.kim\}@snu.ac.kr
}
}

\makeatletter
\let\old@maketitle\@maketitle
\renewcommand{\@maketitle}{%
  \old@maketitle
  \centering
  \includegraphics[
    width=0.85\textwidth,
    trim=10pt 437pt 850pt 29pt,
    clip
  ]{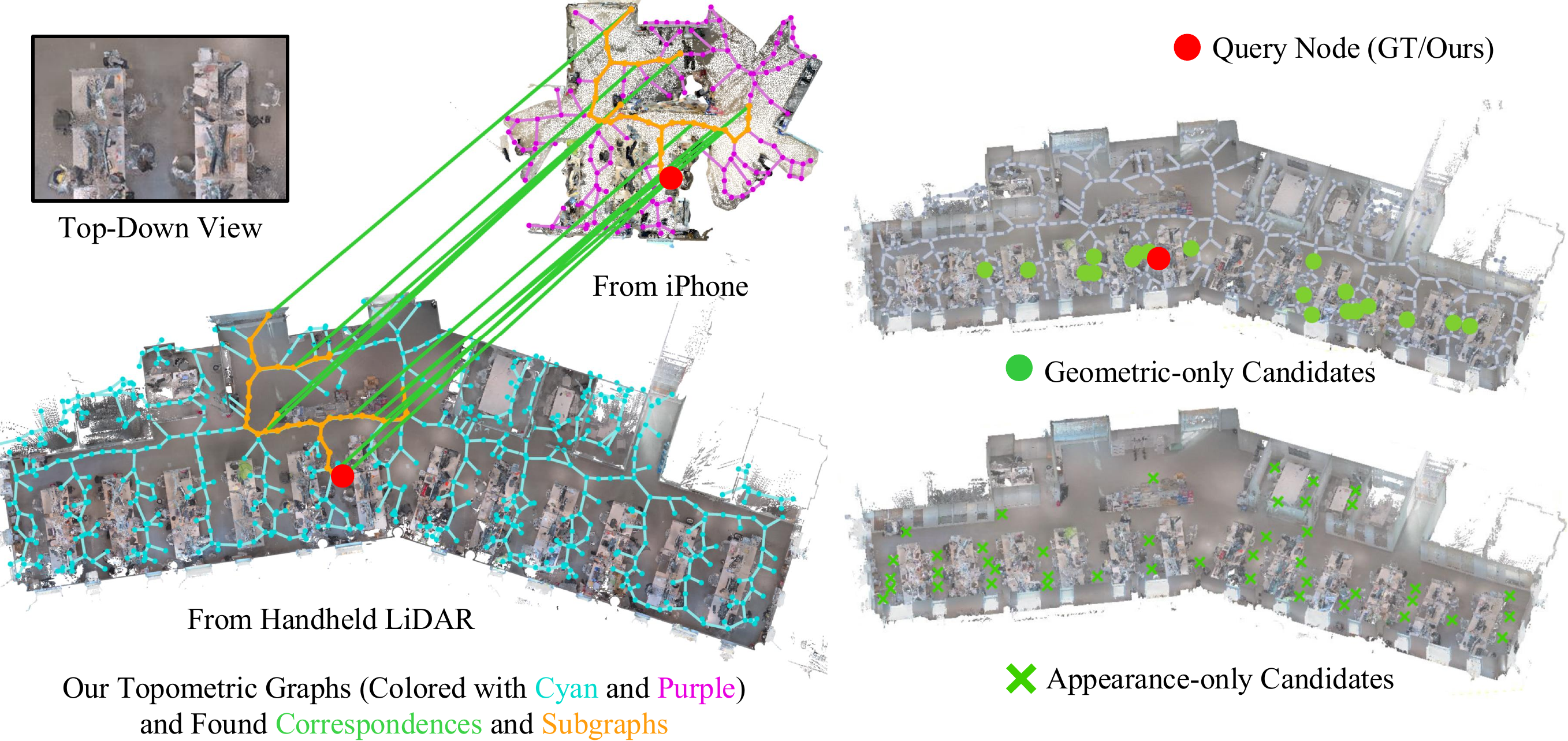}
  \setcounter{figure}{0}
  \captionof{figure}
  {\textbf{\color{wacvcolor}Real-world office scene with repetitive structure captured weeks apart with heterogeneous sensors.} 
   {\color{wacvcolor}3D scans with limited overlap are independently collected using a handheld LiDAR scanner and a mobile phone, respectively, and related via the proposed sparse topometric graphs.}
  For a selected query node, geometry-only candidates shown in green and appearance-only candidates from DINOv2 features~\cite{oquab2023dinov2} remain highly ambiguous. Graph-structured matching resolves this ambiguity and identifies the correct correspondence.}
  \vspace{-2mm}
  \label{fig:main_teaser}
  \par\vspace{7mm}
}
\makeatother

\begin{document}
\maketitle
\begin{abstract}
A conventional embodied agent in an indoor environment builds a map by accumulating noisy real-world measurements and results from heavy inference to guide navigation.
While the detailed measurements may differ due to scene changes or sensor settings, the unique topological structure of navigable passages or the connectivity between rooms may provide training-free guidance that complements local observations. 
We propose a sparse topometric graph as an efficient structural representation for robust indoor navigation, alleviating the need for fine-grained matching, which is computationally expensive, memory-intensive, and prone to real-world variations.
The map consists of nodes that encode local 3D geometry and distance-weighted edges that capture traversable connectivity.
A learned graph-matching network finds consistent subgraph structures across sessions by considering both node-level geometric descriptors and neighborhood connectivity.
The holistic contextual matching is robust to local ambiguities caused by weak distinctiveness or repetitive structure, and establishes stable correspondences despite local scene changes, different sensor modalities, or limited spatial overlap.
We analyze the practicality of the graph through \textit{cross-session point-goal navigation}, where a goal specified in a prior reference map must be reached from an unknown initial pose in a later session.
Experiments show that our method achieves the highest success rates (77.1\% on Gibson~\cite{xiazamirhe2018gibsonenv} and 40.0\% under 3D-FRONT~\cite{fu20213d} scene changes), while reducing memory by more than two orders of magnitude compared to point-level representations. 
The proposed approach further supports cross-session alignment and goal-directed navigation with strong scalability and robust real-world transfer.
\end{abstract}
    
\section{Introduction}
\label{sec:intro}

Embodied agents in daily life need to understand spatial context to interact in unstructured environments.
Capturing the spatial map of the deployed environment often requires composing multiple measurements with minimal overlap, as indoor scenes often contain complex concavities, frequent occlusions, and confined navigable spaces due to furniture, rooms, and narrow passages with corners.
Reusing the acquired map in a separate session poses additional difficulties{\color{wacvcolor}.} 
Transferring the high-quality geometry of the entire deployed space may require extensive spatial memory, and reusing the map with a different embodiment may encounter scene changes and different sensor configurations with unknown relative poses.

Prior navigation studies~\cite{chaplot2020learning, ramakrishnan2020occupancy, savinov2018semi, chaplot2020neural} have used different spatial representations for mapping, localization, and planning in 2D top-down maps of large-scale indoor environments.
Dense metric maps~\cite{chaplot2020learning, ramakrishnan2020occupancy, hornung2013octomap, chaplot2020neural}  preserve detailed geometry and support path planning, but their storage and matching costs rapidly increase with environment scale.
Keyframe-~\cite{campos2021orb, rosinol2020kimera, cadena2016past} and appearance-based~\cite{arandjelovic2016netvlad, keetha2023anyloc, kwon2023renderable, savinov2018semi} maps are compact and visually informative, but their reliability can degrade under viewpoint and scene changes.
Semantic and hierarchical scene graphs~\cite{gu2024conceptgraphs, hughes2022hydra, wald2020learning, yin2024sg} provide higher-level structure through layouts, objects, or language descriptions, but often depend on reliable detections, explicit labels, or predefined semantic categories.

In this paper, we propose a sparse topometric graph that can reliably capture geometric context despite prevalent ambiguities in indoor environments with repetitive structures, moved furniture, and heterogeneous sensor modalities.
In the presence of such variations, local features are insufficient to extract reliable geometric context, and a broader spatial arrangement must be considered.
Our proposed graph representation summarizes the spatial organization of traversable space. {\color{wacvcolor}Each node is centered on traversable structure rather than on a sampled point in the cloud, aggregates the observed 3D points within a fixed-radius neighborhood, and stores a compact geometry-only descriptor, while each edge represents free-space connectivity with a distance-based weight.}
{\color{wacvcolor}This fixed-radius aggregation makes local geometry reusable at the graph level instead of tying descriptors to raw point samples~\cite{qin2022geotransformer,seo2025bufferx}.}
Compressing each local point set into a fixed-size node descriptor keeps the map size proportional to the number of {\color{wacvcolor}graph nodes} rather than the raw scene geometry{\color{wacvcolor}, while preserving enough local and structural information for cross-session map reuse}.

We demonstrate the usability of the representation in \textit{cross-session point-goal navigation}.  
The agent starts from an unknown initial pose and must reach the goal pose, defined on the reference map built in a previous session.
The agent first aligns its current observations relative to the reference map and selects waypoints toward the goal while taking the navigable path topology into account.
We train a graph-matching network to predict cross-session correspondences between graph nodes by jointly considering local geometry and connectivity between neighboring nodes.
The predicted correspondences are used to estimate the relative transformation between the reference and current sessions, after which the graph-based navigation policy guides the agent toward the transferred goal.
{\color{wacvcolor}This setting lets us study whether a single sparse representation can support both cross-session alignment and downstream navigation.}

Fig.~\ref{fig:main_teaser} illustrates our pipeline with a real-world cross-session example.
The example highlights a key challenge. 
Local geometric descriptors or appearance-based cues can be ambiguous when similar structures repeat across partially overlapping sessions.
The proposed representation and graph matching reduce these ambiguities by considering both node-level geometry and neighboring graph connectivity.

The main contributions are summarized as follows:
\begin{itemize}
    \item We propose a sparse topometric graph {\color{wacvcolor}for cross-session indoor map reuse}. It represents traversable space with compact geometry-only node descriptors and distance-weighted graph connectivity, providing a scalable map representation.
    \item {\color{wacvcolor}We introduce a graph-node descriptor that aggregates fixed-radius 3D observations around reusable node centers. This makes local geometry available at the graph level instead of tying descriptors to individual point-cloud samples.}
    \item We introduce a learned graph-matching approach that predicts cross-session node correspondences using both local geometric similarity and graph connectivity, {\color{wacvcolor}and we use the same graph for cross-session alignment and waypoint selection.}
    \item We evaluate the method on simulated cross-session navigation, large-scale indoor scenes, scene-change settings, and {\color{wacvcolor}a real-world indoor study}, demonstrating efficient and reliable cross-session map use.
\end{itemize}

\begin{figure*}[t]
    \centering
    \includegraphics[
    width=\textwidth,
    trim=10pt 490pt 570pt 7pt,
    clip
  ]{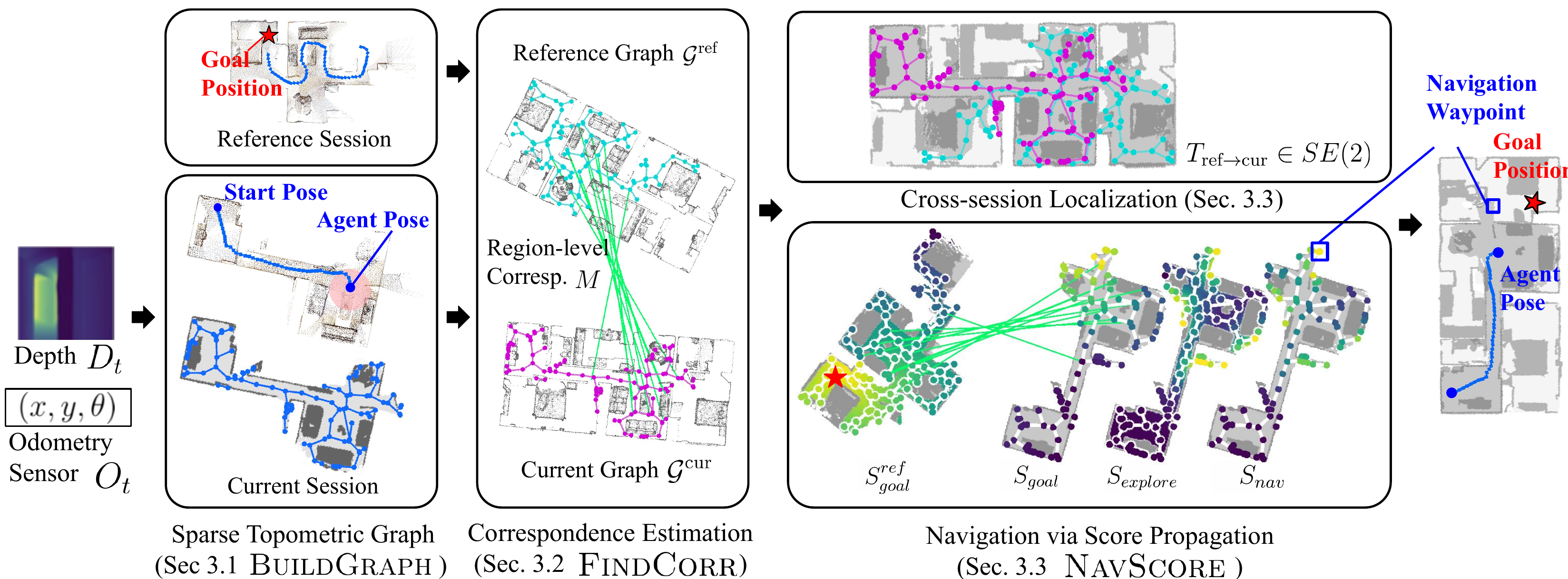}

    \caption{\textbf{
    Overview of the proposed cross-session localization and navigation framework.}
    Given depth $D_t$ and odometry $O_t$ inputs, the agent constructs a topometric graph using medial-axis decomposition and assigns geometric descriptors from local point clouds. Node correspondences $M$ are estimated across reference graph $\mathcal{G}^\text{ref}$ and current graph $\mathcal{G}^\text{cur}$ via graph matching. The correspondences are reused for both localization and navigation. The navigation policy combines the goal-directed score $S_\text{goal}$ with the frontier-based exploration score $S_\text{explore}$ to select a node as the next waypoint toward the goal position $g$ of the cross-session navigation task.
    }
   \vspace{-4mm}

    \label{fig:overview}
\end{figure*}

\section{Related Work}
\label{sec:related_work}

\subsection{Map Representations for Navigation}

\textit{Metric maps}, such as occupancy grids and voxel maps~\cite{ramakrishnan2020occupancy, hornung2013octomap}, provide fine-grained geometric detail for obstacle avoidance and motion planning.
However, their reliance on globally consistent poses makes them vulnerable to accumulated drift and scene changes across sessions, often leading to degraded map quality~\cite{schmid2024khronos}.
\textit{Appearance-based methods}~\cite{kwon2023renderable, shah2023vint} represent scenes via stored images or learned visual features, enabling viewpoint-aware matching and memory-efficient planning. 
Visual place recognition (VPR) approaches~\cite{keetha2023anyloc} extend this to cross-session localization by matching compact global descriptors between sessions.
They are more robust to spatial drift but remain sensitive to lighting, occlusion, and viewpoint deviation.  
\textit{Object-centric scene graphs}~\cite{gu2024conceptgraphs, hughes2022hydra, maggio2024clio, garg2025objectreact} offer higher-level reasoning by representing objects and their relations, supporting semantically meaningful planning.
In particular, ObjReact~\cite{garg2025objectreact} targets cross-session navigation by constructing an object-centric 3D scene graph and reactively planning routes toward recognized goal-relevant objects.
However, these approaches rely on consistent object detection and semantic stability across time.
To support a fair and diverse evaluation, we adopt representative methods from each of the above categories as baselines for cross-session navigation: a metric-based grid map~\cite{chaplot2020learning}, appearance-based maps~\cite{kwon2023renderable,keetha2023anyloc}, and object-centric scene graphs~\cite{gu2024conceptgraphs,garg2025objectreact}.

Beyond these, \textit{point-level geometric descriptors}~\cite{qin2022geotransformer, seo2025bufferx} enable fine-grained spatial alignment and have demonstrated strong performance in localization and map registration tasks.
Rather than using them densely, we adopt their desirable geometric properties in a compact, node-anchored form, combining their discriminative power with structural abstraction for scalable cross-session navigation.

Our work bridges these extremes by introducing a sparse topometric graph that generates a topological graph over navigable areas, where each graph node is anchored with sparse 3D geometric descriptors.
This structure supports both cross-session localization and goal-directed planning in a single pipeline, offering a compact and robust map representation for cross-session navigation.

\subsection{Graph-based Correspondence}
Graph-based representations encode spatial relationships that support structural reasoning for localization and navigation. 
Factor graphs have long provided the backbone for multi-session SLAM~\cite{sh-ch1-fg4slam}, while learned approaches construct topological maps where nodes represent visited locations and edges encode navigability~\cite{chaplot2020neural, savinov2018semi}.
Recent work studies cross-session localization and graph-based alignment over object- or scene-level abstractions, with node-level features drawn from different modalities and semantics~\cite{miao2024scenegraphloc,sarkar2023sgaligner,xie2024sg}.
These approaches demonstrate that reasoning about relational structure helps resolve ambiguities that local features alone cannot resolve.

However, object-level graphs rely on stable semantic detection, which degrades under scene changes or sensing heterogeneity. 
Related work has also explored learned geometric descriptors and registration over point-cloud features~\cite{qin2022geotransformer,seo2025bufferx}.
Such methods typically operate at the point or object level rather than over navigable spatial regions.

Our graph provides a sparse topological description of traversable regions that compactly encodes reachability from the agent's current position.
The graph structure also enables holistic cross-session matching of connected regions, which is robust to perturbations in local features and to moderate topological deviations arising from scene changes or measurement discrepancies.

{\color{wacvcolor}
\begin{algorithm}[t]
\caption{Cross-session navigation with a sparse topometric graph}
\label{alg:pipeline}
\small
\begin{algorithmic}[1]
\Require $\mathcal{G}^{\text{ref}}$, goal $g$, stream $\{(D_t,O_t)\}_{t=1}^{T}$, matcher $\mathrm{GNN}_{\theta}$, update period $K$
\Ensure Next-step action $a_t$
\Function{BuildGraph}{$\{(D_\tau,O_\tau)\}_{\tau=1}^{t}$}
    \State Fuse observations into $\mathcal{O}_t$ and $\mathcal{P}_t$ \label{line:buildgraph:start}
    \State Extract reachable free space from $\mathcal{O}_t$
    \State Sample nodes and connect neighbors
    \State Downsample $\mathcal{P}_t$ and encode radius-based node descriptors
    \State \Return $\mathcal{G}^{\text{cur}}_t, \mathcal{O}_t$ \label{line:buildgraph:end}
\EndFunction
\Function{FindCorr}{$\mathcal{G}^{\text{ref}}, \mathcal{G}^{\text{cur}}$}
    \State $P \gets \mathrm{Sinkhorn}(\mathrm{GNN}_{\theta}(\mathcal{G}^{\text{ref}}, \mathcal{G}^{\text{cur}}))$ \label{line:findcorr:start}
    \State $\mathcal{M} \gets$ argmax matches after removing unmatched assignments from $P$
    \State \Return $\mathcal{M}_{\mathrm{in}} \gets \mathrm{RANSAC}(\mathcal{M})$ \label{line:findcorr:end}
\EndFunction
\Function{NavScore}{$\mathcal{M}_{\mathrm{in}}, \mathcal{G}^{\text{cur}}, g$}
    \If{$\mathcal{M}_{\mathrm{in}} \neq \emptyset$} \label{line:navscore:start}
        \State Estimate relative transform and transfer goal scores
        \State Compute frontier score $\bar{S}_{\mathrm{explore}}$
        \State \Return $\bar{S}_{\mathrm{goal}} + \bar{S}_{\mathrm{explore}}$
    \Else
        \State Compute frontier score $\bar{S}_{\mathrm{explore}}$
        \State \Return $\bar{S}_{\mathrm{explore}}$
    \EndIf \label{line:navscore:end}
\EndFunction
\Procedure{Navigate}{}
    \State Initialize $\mathcal{M}_{\mathrm{in}} \gets \emptyset$ and $v^* \gets \emptyset$
    \State Initialize running map $\mathcal{O}_t$
    \For{$t = 1$ to $T$}
        \State Update $\mathcal{O}_t$ with $(D_t,O_t)$
        \If{$t \bmod K = 0$}
            \State $\mathcal{G}^{\text{cur}}, \mathcal{O}_t \gets$ \Call{BuildGraph}{$\{(D_\tau,O_\tau)\}_{\tau=1}^{t}$}
            \State $\mathcal{M}_{\mathrm{in}} \gets$ \Call{FindCorr}{$\mathcal{G}^{\text{ref}}, \mathcal{G}^{\text{cur}}$}
            \State $S_{\mathrm{nav}} \gets$ \Call{NavScore}{$\mathcal{M}_{\mathrm{in}}, \mathcal{G}^{\text{cur}}, g$}
            \State $v^* \gets \arg\max_v S_{\mathrm{nav}}(v)$
        \EndIf
        \State $a_t \gets \mathrm{PathPlanner}(\mathbf{p}_{v^*}, \mathcal{O}_t)$
    \EndFor
\EndProcedure
\end{algorithmic}
\end{algorithm}
}

\section{Method}
\label{sec:method}

We introduce a sparse topometric graph representation and a pipeline for cross-session localization and navigation.
We consider a cross-session navigation task in which an agent operates in an indoor environment for which a reference map was constructed in a prior session.
The reference session is represented as a sparse topometric graph $\mathcal{G}^{\text{ref}}$.
At deployment, the agent is at an unknown pose within the same environment and must navigate to a goal position $g \in \mathbb{R}^2$ specified in the coordinate frame of the reference session.
{\color{wacvcolor}The current session may differ from the reference due to trajectory variation, partial spatial overlap, accumulated odometry drift, and local scene arrangement.}

An overview of the proposed approach is shown in Fig.~\ref{fig:overview}, and Alg.~\ref{alg:pipeline} summarizes the deployment loop.
%
At each time step $t$, the agent receives a depth observation $D_t$ and an odometry measurement $O_t = (x_t, y_t, \theta_t)$ in its local frame.
{\color{wacvcolor}Using the stream $\{(D_\tau,O_\tau)\}_{\tau=1}^{t}$, the agent incrementally constructs a top-down occupancy map $\mathcal{O}_t$ and a depth-derived accumulated point cloud $\mathcal{P}_t$. 
Every $K{=}20$ steps, these geometric observations are converted into a current-session graph $\mathcal{G}^{\text{cur}}_t$. (\textsc{BuildGraph}, Sec.~\ref{subsec:map_representation})}.
To enable cross-session localization, the agent performs topometric graph matching between $\mathcal{G}^{\text{cur}}_t$ and $\mathcal{G}^{\text{ref}}$, producing candidate node correspondences $\mathcal{M}$ (\textsc{FindCorr}, Sec.~\ref{subsec:graph_matching}).
These correspondences are then used for relative pose estimation and navigation toward the reference-session goal, without relying on dense map alignment or continuous pose tracking across sessions (\textsc{NavScore}, Sec.~\ref{subsec:localization}).

\subsection{Sparse Topometric Graph}
\label{subsec:map_representation}

{\color{wacvcolor}The \textsc{BuildGraph} stage in Alg.~\ref{alg:pipeline} (lines~\ref{line:buildgraph:start}--\ref{line:buildgraph:end}) constructs the shared representation used by the rest of the pipeline.}
We represent each session as a \textit{topometric} graph $\mathcal{G} = (\mathcal{V}, \mathcal{E})$.
Here, \textit{topometric} denotes the joint encoding of \textit{topological connectivity} among navigable anchors and \textit{metric} and \textit{geometric} information.
{\color{wacvcolor}Each node $v_i \in \mathcal{V}$ corresponds to a navigable anchor extracted from traversable free space and is associated during graph construction with a 2D position $\mathbf{p}_i \in \mathbb{R}^2$ and a sparse 3D geometric descriptor $f_i \in \mathbb{R}^{32}$.}
Edges $\mathcal{E}$ encode traversable connectivity between adjacent nodes, where each edge weight stores the local distance $d_{ij} = \lVert \mathbf{p}_i - \mathbf{p}_j \rVert_2$ between neighboring anchors.

The graph is derived from two complementary geometric representations.
{\color{wacvcolor}A top-down occupancy map is used to identify reachable traversable free space and extract graph nodes, while a depth-derived accumulated point cloud is used to compute geometric descriptors within each node.}
We first extract the portion of free space that is reachable from the agent's current position so that the graph remains focused on reusable navigable structure rather than on every nominally free cell in the dense map.
{\color{wacvcolor}Navigable anchors are defined by the structure of the reachable free-space occupancy map, with graph nodes placed at uniform intervals of $d{=}5$ cells ($0.25$\,m) along each branch.
The nodes anchor local traversable areas from which features are extracted.}

The key geometric design choice is how node descriptors are defined.
{\color{wacvcolor}Unlike the original per-point formulation, descriptors are anchored at navigable graph nodes rather than at existing points in the cloud.}
For each node $v_i$, we collect points $\{q \in \mathcal{P} : \|q - \mathbf{p}_i\| < r\}$ within a fixed radius around its construction anchor and compute a local geometric embedding.
We adopt a pre-trained rotation-equivariant point-cloud encoder from BUFFER-X~\cite{seo2025bufferx} with frozen weights, providing robustness to partial observations and sensor heterogeneity.
{\color{wacvcolor}As a result, the representation remains compact while each node still carries both local 3D context and graph-level connectivity for cross-session correspondence and downstream navigation. Implementation details such as voxel downsampling are deferred to the supplementary material.}

\begin{figure}[t]
  \centering
    \includegraphics[
    width=\columnwidth,
    trim=80pt 30pt 380pt 100pt,
    clip
  ]{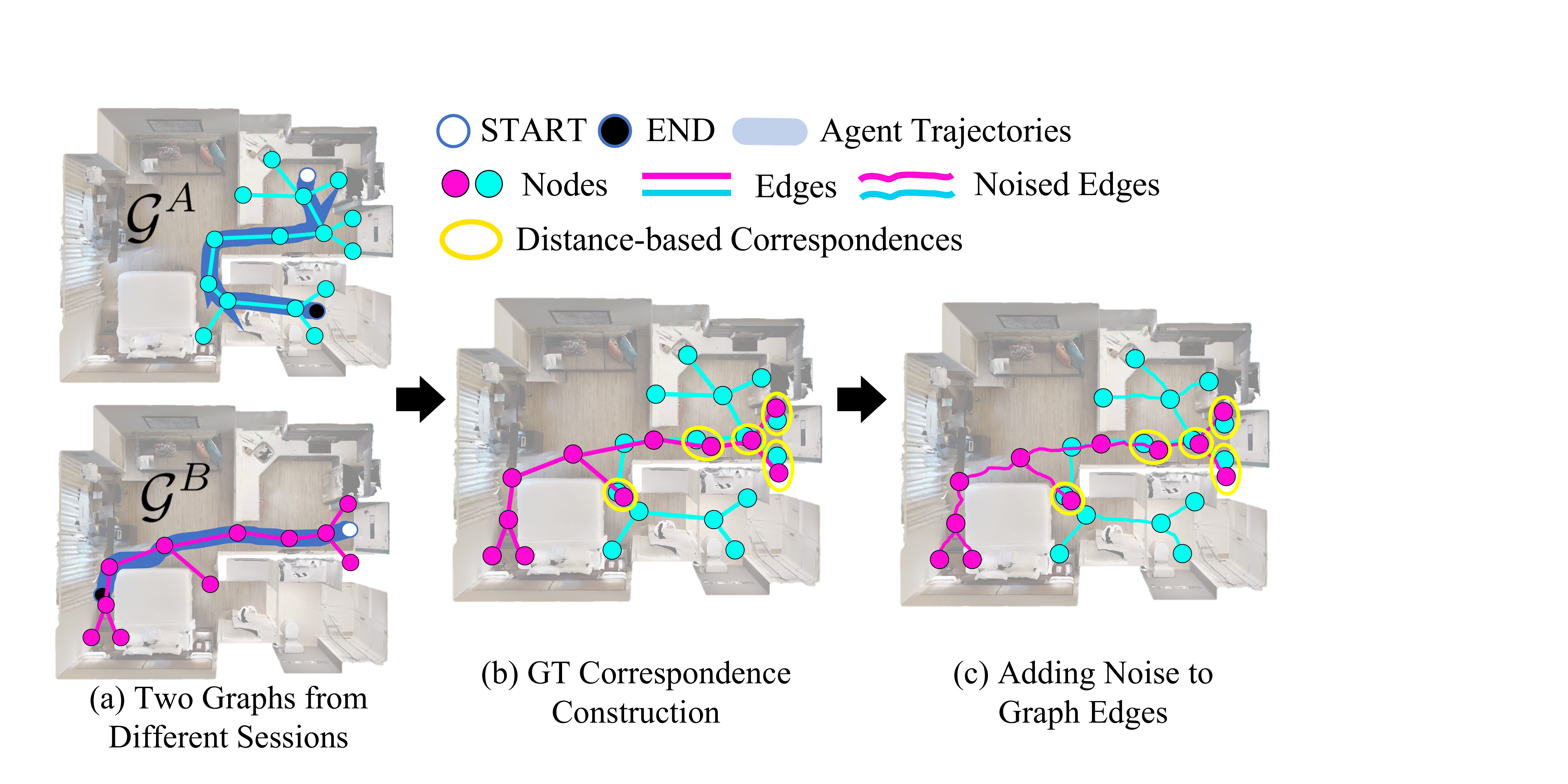}
  \caption{
    \textbf{Training pair construction for graph matching.}
    (a) Topometric graphs are extracted from two sessions with partially overlapping coverage of the same scene.
    (b) Ground-truth node correspondences are established by pairwise Euclidean distances between node centers across sessions.
    (c) Gaussian noise is injected into the edge weight to simulate odometry drift and construct training graph pairs. 
}
\vspace{-2mm}
  \label{fig:graph_corr}
\end{figure}

\subsection{Correspondence Estimation}
\label{subsec:graph_matching}

{\color{wacvcolor}The \textsc{FindCorr} stage in Alg.~\ref{alg:pipeline} (lines~\ref{line:findcorr:start}--\ref{line:findcorr:end}) estimates cross-session correspondences.}
Given topometric graphs constructed from different navigation sessions, our goal is to identify correspondences between nodes representing the same physical locations in the environment.
This problem requires reasoning over a noisy topometric graph, in which both the local geometry encoded in nodes and the graph structure are perturbed by measurement noise, odometry drift, and scene variation.
We treat the topometric graph as a shared representation, from which node correspondences are inferred as a signal used jointly for localization and navigation.
These correspondences are \textit{reused} for both \textit{pose estimation} and \textit{goal-directed navigation}.

Rather than relying on dense frame- or point-level matching, which can be disrupted by unobserved areas or local scene variations, our approach operates on sparse graphs.
Each node aggregates local 3D geometry over a traversable anchor using rotation-robust and view-invariant descriptors (Sec.~\ref{subsec:map_representation}), yielding a compact representation that naturally tolerates missing coverage under partial overlap and sensing heterogeneity.
In addition, the graph structure encodes spatial relationships among nodes, providing global context that helps disambiguate correspondences beyond local geometry.

\noindent\textbf{Matching Objective.}
{\color{wacvcolor}Our objective is to predict a soft partial assignment between nodes in the reference and current graphs by jointly reasoning over node-level geometric features and graph connectivity.
We employ IPCA-GM~\cite{WangPAMI20} with Sinkhorn-normalized assignment to estimate correspondence probabilities while allowing unmatched nodes under partial overlap. 
Because the prediction is constrained by both local geometry and graph structure, the resulting correspondences are more robust to local ambiguity, repeated structure, and incomplete coverage than geometry-only matching.}


\noindent\textbf{Training Pair Construction.}
{\color{wacvcolor}As illustrated in Fig.~\ref{fig:graph_corr}, training pairs are constructed from two partial traversals collected in the same scene with different spatial coverage. 
We build graphs $\mathcal{G}^{A}$ and $\mathcal{G}^{B}$ from these sessions and define correspondence supervision by spatial proximity between their construction anchors across sessions. 
During training, we perturb traversable edge distances to improve robustness to odometry drift. 
Exact objective and implementation details are deferred to the supplementary material.}



\subsection{Localization and Waypoint Selection}
\label{subsec:localization}

{\color{wacvcolor}The \textsc{NavScore} stage in Alg.~\ref{alg:pipeline} (lines~\ref{line:navscore:start}--\ref{line:navscore:end}) reuses the predicted correspondences for both localization and waypoint selection.}
Using the RANSAC-filtered node matches $\mathcal{M}_{\mathrm{in}}$, each matched node pair provides a position correspondence $(\mathbf{p}_i^{\text{ref}}, \mathbf{p}_j^{\text{cur}})$.
We estimate the relative transformation $T_{\text{ref}\rightarrow\text{cur}} \in SE(2)$ from these inlier correspondences.
During deployment, odometry drift accumulates over time as the agent explores.
{\color{wacvcolor}To reduce its impact on localization, we construct the current-session graph from a spatially local subset of recently observed traversable structure rather than from the full accumulated map.}

Our navigation policy directly reuses the topometric graph and cross-session correspondences introduced earlier, without additional representations or task-specific supervision.
Given the correspondence, we perform cross-session navigation by selecting nodes in the current graph as waypoints.
The policy balances exploitation of available goal information and exploration of unmapped areas using two complementary signals, a goal-directed score $S_{\text{goal}}$ derived from cross-session correspondences and a frontier-based exploration score $S_{\text{explore}}$ that promotes explored area expansion.
{\color{wacvcolor}A goal $g \in \mathbb{R}^2$ specified in the reference-session frame is converted into scores on the reference graph, and those scores are propagated to the current-session graph through the predicted correspondence. The exploration term is computed on the current occupancy map and evaluated at current-session graph nodes for waypoint selection.}
Since the correspondence matches the holistic spatial context rather than individual nodes, reliable cross-session evidence becomes available only after sufficient exploration.
{\color{wacvcolor}We therefore transition from using the exploration score only to the combined goal-directed and exploration score once cross-session evidence becomes reliable, as illustrated in Fig.~\ref{fig:overview}. 
The selected waypoint is then executed on the occupancy map using a Fast Marching Method planner~\cite{sethian1996fast}.}

\section{Experiments}
\label{sec:experiments}

\begin{table}[t]
\centering
\caption{Evaluation protocol across datasets.}
\label{tab:eval_protocol}
\small
\setlength{\tabcolsep}{4pt}
\begin{tabular}{lccc}
\toprule
Dataset & Max Steps & Goal Thr. (m) & Avg. Area (m$^2$) \\
\midrule
3D-FRONT & 500 & 1.0 & 80.6 \\
Gibson & 500 & 2.0 & 83.4 \\
HM3D & 1000 & 2.0 & 212.8 \\
\bottomrule
\end{tabular}
\end{table}

We evaluate our approach on cross-session point-goal navigation experiments.
We assess navigation under odometry noise, varying scene scales, limited inter-session overlap, and scene change. 

\subsection{Experimental Setup}

We evaluate on three datasets: Gibson~\cite{xiazamirhe2018gibsonenv} for the main evaluation under noiseless and noisy odometry, HM3D~\cite{ramakrishnan2021habitat} for large-scale evaluation, and 3D-FRONT~\cite{fu20213d} for scene-change evaluation.

Experiments are conducted in Habitat~\cite{habitat19iccv} \textcolor{wacvcolor}{using the LoCoBot action space~\cite{locobot}. The agent takes discrete actions of moving forward by 0.25\,m or turning left or right by $10^\circ$. The success threshold is 2.0\,m for Gibson and HM3D. For 3D-FRONT, we use a 1.0\,m threshold because the primary question is whether the representation remains locally reusable under scene change, rather than long-range planning. Table~\ref{tab:eval_protocol} summarizes the step budgets and scene scales used in each dataset.} The graph matching network is trained on Gibson training scenes and is applied without retraining across all datasets. \textcolor{wacvcolor}{Unless otherwise noted, all methods are executed under the same simulator and action setting. Additional implementation details are deferred to the supplementary material.}

\textcolor{wacvcolor}{In every episode, the goal is specified on the reference-session map, while the agent acts only in the current session. The reference and current sessions are collected at different times and from different start positions. The agent therefore cannot rely on frame-to-frame continuation across sessions and must recover a usable cross-session relation from partial overlap. This shared protocol lets us compare how well each representation supports map reuse rather than single-session exploration.}

\subsection{Baselines}

We compare against \textcolor{wacvcolor}{representative baselines spanning several reusable map abstractions}. ANS~\cite{chaplot2020learning} represents the environment with a dense occupancy-grid map. RnR-Map~\cite{kwon2023renderable} uses a top-down visual feature map. \textcolor{wacvcolor}{AnyLoc~\cite{keetha2023anyloc} provides an appearance-based retrieval baseline using compact global descriptors. ConceptGraph~\cite{gu2024conceptgraphs} and} ObjectReact~\cite{garg2025objectreact} represent the environment with \textcolor{wacvcolor}{scene-graph or} object-graph abstractions. \textcolor{wacvcolor}{We also compare against BUFFER-X~\cite{seo2025bufferx} in the compactness analysis as a dense point-level geometric reference.}

\textcolor{wacvcolor}{We prioritize baselines that expose an explicit reference-session representation and can be integrated into a goal-conditioned navigation loop in the current session. This lets us compare dense occupancy maps, top-down feature maps, appearance-based retrieval, and semantic graph abstractions under the same cross-session point-goal protocol. Several baselines were not originally proposed for this exact setting, so we adapt them while preserving their original representation and decision logic as much as possible. Full adaptation details are deferred to the supplementary material.}

\begin{table}[t]
\centering

\caption{Cross-session point-goal navigation results on Gibson~\cite{xiazamirhe2018gibsonenv} under noiseless and noisy odometry.}
\label{tab:gibson_nav}
\begin{tabular}{lcccccc}
\toprule
 & \multicolumn{2}{c}{Noiseless} & \multicolumn{2}{c}{Noisy ($\sigma=1.0$)} \\
\cmidrule(lr){2-3} \cmidrule(lr){4-5}
Method & SR & SPL & SR & SPL \\
\midrule
ANS~\cite{chaplot2020learning} & 40.00 & 33.40 & 40.00 & {33.13} \\
RnR~\cite{kwon2023renderable} & 30.00 & 23.67 & 41.41 & 32.38 \\
AnyLoc~\cite{keetha2023anyloc} & 40.00 & 10.91 & 30.00 & 5.69 \\
ObjReact~\cite{garg2025objectreact} & 54.28 & 40.81 & 54.28 & \textbf{40.81} \\
ConceptGraph~\cite{gu2024conceptgraphs} & 41.43 & 22.93 & 32.86 & 16.40 \\
Ours & \textbf{77.14} & \textbf{47.71} & \textbf{60.00} & 23.64 \\
\bottomrule
\end{tabular}
\end{table}

\subsection{Navigation Performance}

\noindent\textbf{Main evaluation on Gibson.}
Table~\ref{tab:gibson_nav} reports cross-session point-goal navigation under noiseless and noisy odometry.
We use Success Rate (SR) and Success weighted by Path Length (SPL)~\cite{anderson2018evaluation}. \textcolor{wacvcolor}{For ObjectReact, we report the result obtained with its original policy-coupled evaluation setting. 
That method uses an RGB-frame-based controller~\cite{shah2022gnm} rather than the same discrete 0.25\,m forward action used by the other methods here, so its SPL is informative but not strictly action-level comparable.} Our method achieves the highest SR under both odometry conditions, and it also achieves the best SPL under noiseless odometry. \textcolor{wacvcolor}{Under noisy odometry, SR remains strong while SPL decreases more noticeably, suggesting that the representation still supports reliable cross-session recovery while path efficiency is more affected by delayed correspondence under drift.}

      

\begin{figure}[t]
  \centering
      \includegraphics[
    width=0.9\columnwidth,
    trim=0pt 180pt 320pt 5pt,
    clip
  ]{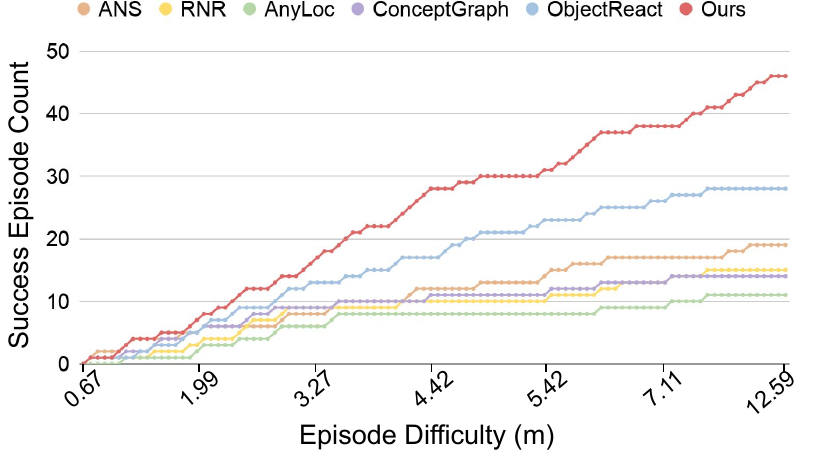}
  \caption{
    \textbf{Navigation performance in large-scale HM3D environments~\cite{ramakrishnan2021habitat}.}
    Cumulative number of successful episodes out of 100 as the geodesic separation between the reference- and current-session start positions increases.
  }
  \label{fig:nav_hm3d}
\end{figure}

\begin{figure*}[t]
  \centering
  \includegraphics[
    width=0.88\textwidth,
    trim=0pt 800pt 660pt 0pt,
    clip
  ]{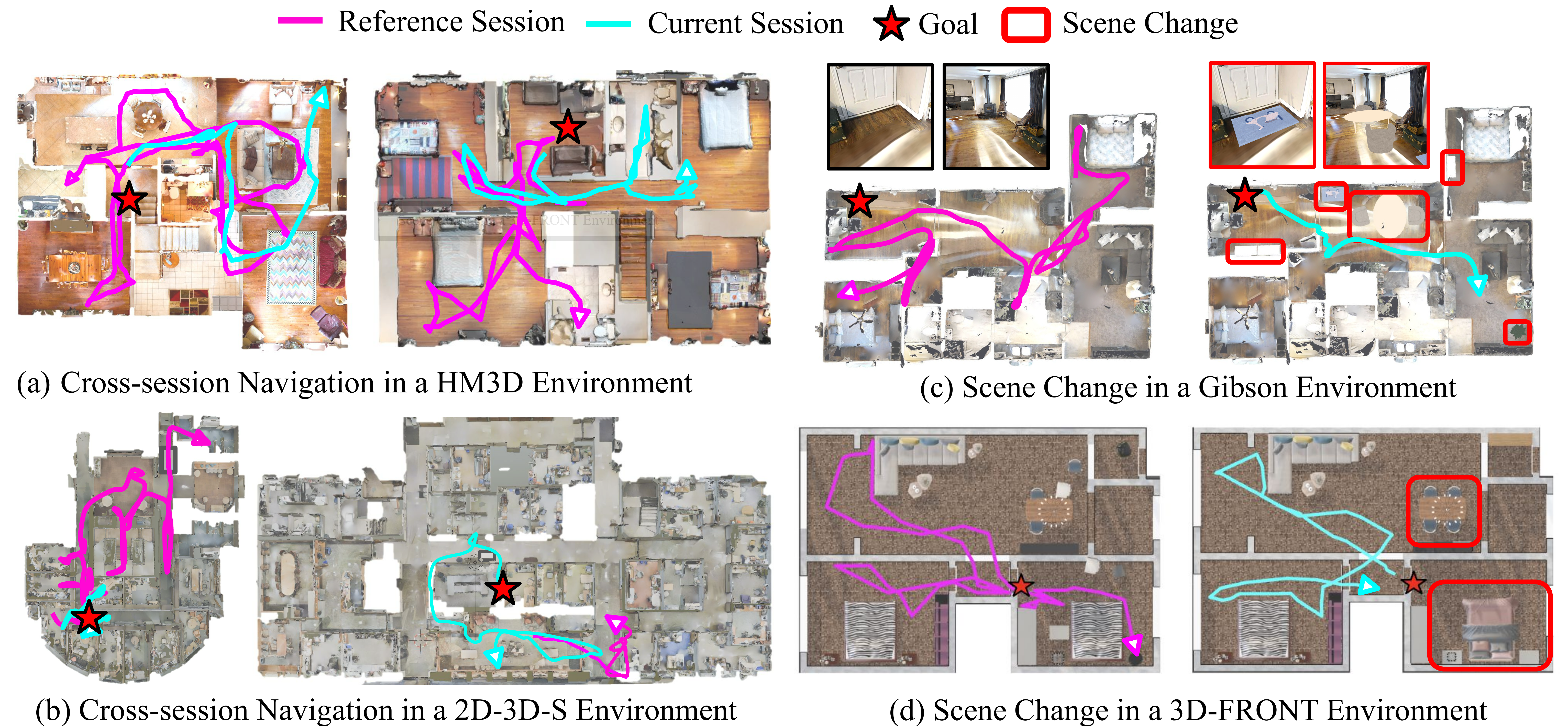}
  \caption{
    \textbf{Qualitative cross-session navigation results.}
    (a–b) Navigation examples in HM3D and 2D-3D-S environments, respectively.
    (c–d) Examples under scene changes in Gibson and 3D-FRONT, where modified regions are highlighted by red boxes. Across these settings, the proposed topometric graph supports cross-session localization and goal-directed navigation despite partial overlap and local environmental changes.
  }
  \label{fig:vis_all}
  \vspace{-2mm}
\end{figure*}

\noindent\textbf{Large-scale evaluation on HM3D.}
Fig.~\ref{fig:nav_hm3d} evaluates navigation as the difficulty of cross-session reuse increases. 
\textcolor{wacvcolor}{For each episode, we first sample a reference-session start position and then choose the current-session start position to be a prescribed geodesic distance away. 
We use this start-position separation as a controlled difficulty proxy. Larger separation generally makes useful overlap harder to recover early in the episode and therefore makes cross-session correspondence more difficult.} 
The figure reports accumulated successful episodes as this difficulty increases. 
\textcolor{wacvcolor}{Our method degrades more gradually than the baselines and maintains the strongest performance in the harder regimes.}

\begin{table}[t]
\centering

\caption{Cross-session point-goal navigation results under scene changes on 3D-FRONT~\cite{fu20213d}.}

\label{tab:scene_change}
\small
\setlength{\tabcolsep}{2pt}
\begin{tabular}{lcccccc}
\toprule
 & ANS & RnR & AnyLoc & ObjReact & ConceptGraph & \textbf{Ours} \\
\midrule
SR  & 13.33 & 13.33 & 20.00 & 6.67 & 13.33 & \textbf{40.00} \\
SPL & 8.36 & 8.83  & 6.91  & 6.67 & 4.92  & \textbf{20.18} \\
\bottomrule
\end{tabular}

\end{table}

\noindent\textbf{Evaluation under scene change.}
Table~\ref{tab:scene_change} reports results on 3D-FRONT with \textcolor{wacvcolor}{furniture additions, removals, rearrangements, and same-category replacements across sessions. 
We use a 1.0\,m success threshold in this setting because the goal is to test whether the representation remains locally reusable when object layout changes.} 
Our method achieves the highest SR and SPL in this setting.
\textcolor{wacvcolor}{This result suggests that the representation remains useful even when object arrangements differ across sessions.} Fig.~\ref{fig:vis_all} shows a qualitative example.
\textcolor{wacvcolor}{The same figure also includes an additional qualitative example from 2D-3D-S~\cite{armeni2017joint} to illustrate behavior in a distinct large indoor environment with repetitive structure.}

\subsection{Compactness}

\begin{table}[t]
\centering
\caption{Memory comparison across methods on Gibson~\cite{xiazamirhe2018gibsonenv}.}
\label{tab:efficiency}
\setlength{\tabcolsep}{2.0 pt}
\resizebox{\columnwidth}{!}{
\begin{tabular}{lcc}
\toprule
Method & Scene Representation & Storage (MB) \\
\midrule
ANS~\cite{chaplot2020learning}
& Grid Map
& 1.76 \\

RnR-Map~\cite{kwon2023renderable}
& Feature Map
& 2.00 \\

AnyLoc~\cite{keetha2023anyloc}
& Place Descriptor Map
& 13.25 \\

ObjReact~\cite{garg2025objectreact}
& Object Graph
& 0.80 \\

ConceptGraph~\cite{gu2024conceptgraphs}
& Scene Graph with Language Features
& 5.00 \\

BUFFER-X~\cite{seo2025bufferx}
& Point Cloud
& 2.81 \\

Ours
& Sparse Topometric Graph
& 0.02 \\
\bottomrule
\end{tabular}
}
\end{table}

\noindent\textbf{Compactness.}
Table~\ref{tab:efficiency} compares the stored reference-session representation reused by each method in the current session. 
Storage is measured on the representation retained after the reference-session traversal under the same episode data budget, rather than on transient intermediate states during construction. 
For example, BUFFER-X stores the reference session as the voxel-downsampled point cloud used as input to matching.
The proposed representation is substantially smaller than dense occupancy grids, dense feature maps, and point-level geometry. 
This follows from storing geometry only at sparse graph-node anchors and traversable connectivity, rather than across the full scene.
Detailed measurement protocol is provided in the supplementary material.

\subsection{Real-World Data Demonstration}
\label{sec:real-world}

We further evaluate the representation on \textcolor{wacvcolor}{real-world cross-session data captured several weeks apart with heterogeneous sensors}. One session uses \textcolor{wacvcolor}{a handheld LiDAR scanner and the other uses a mobile phone with built-in LiDAR}. The two sessions are collected independently at different times of day. \textcolor{wacvcolor}{The LiDAR session was recorded in daytime, and the mobile-phone session was recorded at night. They cover only partially overlapping portions of the environment.} The scene is an office building with repeated monitors, tables, and cubicles of similar design, which creates strong local ambiguity in both geometry and appearance across nearby locations. \textcolor{wacvcolor}{We use this example as a focused correspondence study rather than a closed-loop robot experiment. Acquisition and reconstruction details are deferred to the supplementary material.}

Fig.~\ref{fig:main_teaser} illustrates the main conceptual value of the representation. The left side shows how the two sessions are related at the graph level through node correspondences. On the right, we select one query node and compare graph-structured matching against two weaker cues in isolation. The first is geometry-only matching based on node-descriptor similarity. \textcolor{wacvcolor}{The second is appearance-only matching based on DINOv2 features~\cite{oquab2023dinov2}.} In this repeated office layout, both isolated cues yield multiple plausible candidates, while the graph-structured representation resolves the ambiguity and identifies the correct correspondence. \textcolor{wacvcolor}{We use this example as supporting evidence for the representation claim of the paper.}

\section{Limitations and Scope}
\label{sec:limitations}

This paper studies \textcolor{wacvcolor}{cross-session map reuse for} single-floor indoor navigation \textcolor{wacvcolor}{under a 2D top-down point-goal formulation}. Multi-floor environments and explicit vertical connectivity are outside the current evaluation scope, and extending the representation to such settings is a natural direction for future work.
Within that scope, we focus on \textcolor{wacvcolor}{cross-session reuse under partial overlap, sensing variation, and moderate scene change}. More severe changes to traversable connectivity, such as closed doors, blocked corridors, or large structural reconfiguration, were not systematically evaluated here and remain an important area for future study.
Finally, the real-world evaluation in this paper is \textcolor{wacvcolor}{a demonstration on independently captured data rather than a full onboard deployment with live control, sensing, and recovery behaviors}. This is still useful for validating the representation, but it also sets a clear boundary on what the current experiments prove.

\section{Conclusion}
\label{sec:conclusion}

We present a sparse topometric graph representation that serves as an effective abstraction for cross-session indoor localization and goal-directed navigation.
Experiments show reliable performance under odometry noise, limited overlap, large-scale scenes, and environmental changes, underscoring the practicality of the representation for cross-session map reuse.

{
    \small
    \bibliographystyle{ieeenat_fullname}
    \bibliography{main}
}

\end{document}